\documentclass[sigconf]{acmart}

\AtBeginDocument{%
  \providecommand\BibTeX{{%
    \normalfont B\kern-0.5em{\scshape i\kern-0.25em b}\kern-0.8em\TeX}}}

\usepackage{multirow}
\usepackage{graphicx}

\usepackage{tikz}
\newcommand*\circled[1]{\tikz[baseline=(char.base)]{
            \node[shape=circle,draw,fill=black,text=white,font=\sffamily\bfseries\small, inner sep=1.2pt] (char) {#1};}}

\newcommand*{\rom}[1]{\expandafter\@slowromancap\romannumeral #1@}

\setcopyright{acmcopyright}
\copyrightyear{2018}
\acmYear{2018}
\acmDOI{10.1145/1122445.1122456}

\begin{document}

\title{AdaSplit: Adaptive Trade-offs for Resource-constrained Distributed Deep Learning}

\author{Ayush Chopra}
\affiliation{%
  \institution{MIT}
  \country{Cambridge, MA}}
 
\author{Surya Kant Sahu}
\affiliation{%
  \institution{BIT}
  \country{Durg, India}}

\author{Abhishek Singh}
\affiliation{%
  \institution{MIT}
  \country{Cambridge, MA}}

\author{Abhinav Java}
\affiliation{%
  \institution{DTU}
  \country{Delhi, India}}
  
\author{Praneeth Vepakomma}
\affiliation{%
  \institution{MIT}
  \country{Cambridge, MA}}
  
\author{Vivek Sharma}
\affiliation{%
  \institution{MIT}
  \country{Cambridge, MA}}
  
\author{Ramesh Raskar}
\affiliation{%
  \institution{MIT}
  \country{Cambridge, MA}
  }

\begin{abstract}
Distributed deep learning frameworks like federated learning (FL) and its variants are enabling personalized experiences across a wide range of web clients and mobile/IoT devices. However, these FL-based frameworks are constrained by computational resources at clients due to the exploding growth of model parameters (eg. billion parameter model). 
Split learning (SL), a recent framework, reduces client compute load by \textit{splitting} the model training between client and server. This flexibility is extremely useful for low-compute setups but is often achieved at cost of increase in bandwidth consumption and may result in sub-optimal convergence, especially when client data is heterogeneous.
In this work, we introduce \textit{AdaSplit} which enables efficiently scaling SL to low resource scenarios by reducing bandwidth consumption and improving performance across heterogeneous clients.
To capture and benchmark this multi-dimensional nature of distributed deep learning, we also introduce \textit{C3-Score}, a metric to evaluate performance under resource budgets.
We validate the effectiveness of \textit{AdaSplit} under limited resources through extensive experimental comparison with strong federated and split learning baselines. We also present a sensitivity analysis of key design choices in AdaSplit which validates the ability of \textit{AdaSplit} to provide adaptive trade-offs across variable resource budgets.
\end{abstract}

\maketitle

\section{Introduction}
\label{sec:intro}
Distributed machine (deep) learning is characterized by a setting where many clients (web browsers, mobile/IoT devices) collaboratively train a model under the orchestration of a central server (eg. service provider), while keeping the training data decentralized. 
As strict regulations emerge for data capture and storage such as GDPR ~\cite{gdpr}, CCPA ~\cite{ccpa}, distributed deep learning is being used to enable privacy-aware personalization across a wide range of web clients and smart edge devices with varying resource constraints. For instance, distributed deep learning is replacing third-party cookies in the chrome \textit{browser} for ad-personalization ~\cite{epasto2021massively,charles2021large}, enabling next-word prediction on \textit{mobile devices} ~\cite{next-word-fl}, speaker verification on \textit{smart home assistants} ~\cite{alexa-speech}, HIPPA-compliant diagnosis on \textit{clinical devices} ~\cite{fl-health-intro} and real-time navigation in \textit{vehicles} ~\cite{fl-self-driving}.

\begin{figure}
    \centering
    \includegraphics[scale=0.34, bb=10 10 710 290]{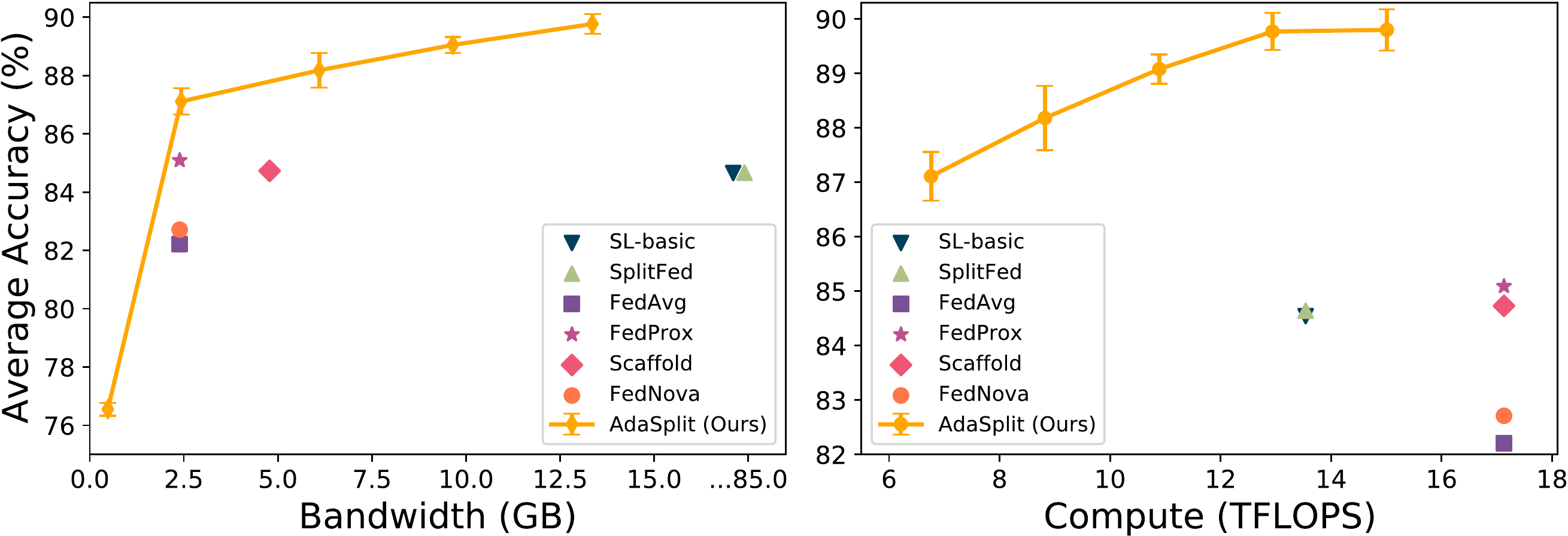}
    \caption{AdaSplit achieves improved accuracy under limited resources (bandwidth $\&$ compute) and can adapt to variable resource budgets. Results on \textit{Mixed-NonIID} dataset.}
    \label{fig:intro-figure}
\end{figure}

A general distributed deep learning pipeline involves multiple rounds of \textit{training} and \textit{synchronization} steps where, in each round, a model is \textit{trained} with local client data and updates across multiple clients are \textit{synchronized} by the server into a global model. Techniques have been proposed with the goal to maximize accuracy under constraints on resource (bandwidth, compute) consumption. Figure ~\ref{fig:intro-figure} compares our proposed \textit{AdaSplit} (in \textcolor{yellow}{yellow}) with strong baselines ~\cite{fedavg, fedprox, scaffold, fednova,gupta2018distributed, thapa2020splitfed} along these dimensions.

\begin{figure*}
    \centering
    \includegraphics[width=0.9\linewidth]{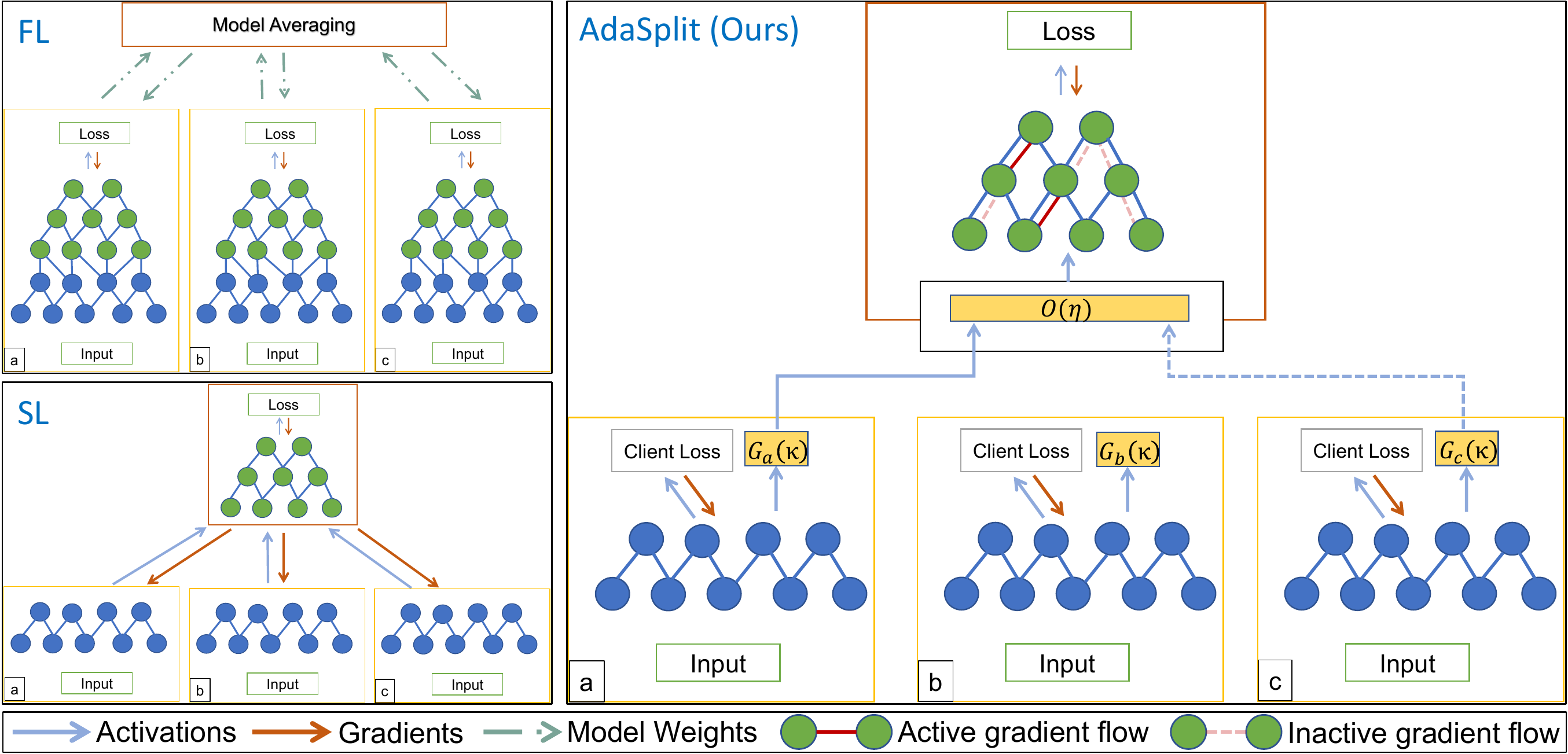}
    \vspace{-0.2cm}
    \caption{Training protocols with N=3 clients for federated learning (FL), split learning (SL) and our proposed AdaSplit which builds upon split learning framework. \textit{AdaSplit} improves i) \textit{\underline{Computation}} using the local client gradient (with $L_{client}$) and training the server intermittently (using gate $G(.)$ parameterized by $\kappa$), ii) \textit{\underline{Communication}} by reducing payload size (no gradient flow from server-client) and interaction frequency (using $O(.)$ parameterized by $\eta$) and iii) \textit{\underline{Collaboration}} by allowing each client to update sparse partition of server parameters (on edges with active gradient flow). Specifically, in this figure, client (b) is in local phase and client (a,c) are in global phase. Client (a) is selected to train and it only updates a sparse partition of server model parameters corresponding to edges with active gradients on the server. The protocol is detailed in Section~\ref{sec:method}. \vspace{-0.5cm}}
    \label{fig:overall-arch}
\end{figure*}

Federated learning (FL) ~\cite{fedavg} is one of the widely studied frameworks ~\cite{fedavg, fedprox, fednova, he2020fedml, cheng2017survey}. In each round of FL, \textit{first}, all clients train a copy of the model locally on their device for several iterations and \textit{then}, communicate the final model parameters with the server which \textit{synchronizes} updates across clients by 
\textit{averaging} all clients model parameters and shares back the unified global model for next training round. This is summarized in Figure ~\ref{fig:overall-arch}. With entire model training on-client, federated learning is challenged by the \textit{compute budgets} of client devices. \textit{First}, on-client model training needs resource-intensive clients (with high-performance GPUs to avoid stragglers) and is increasingly becoming impractical due to exploding growth in model sizes (eg. billion parameter models for language and image modeling  ~\cite{gpt, bert, vit}). \textit{Second}, as the number of clients (and/or model sizes) scales, bandwidth requirements for the system may worsen as entire models need to be communicated between client and server. \textit{Furthermore}, at inference, often, storing the entire trained model on-client can have intellectual property implications that limit real-world usability.

More recently, the split learning (SL) framework ~\cite{gupta2018distributed, thapa2020splitfed, poirot2019split, splitlearning2, vepakomma2018split} has emerged to alleviate some of these concerns in federated learning. SL reduces client computation load by involving the server in the training process.
In each round, clients take turns to interact with the server for multiple iterations where they update parameters of a local model on the client and a (shared) global model residing on the server. Specifically, at each iteration, the client model generates input activations that are communicated to the server which uses them to compute gradients that are used to train the server model and transmitted to the client to train the client model.
This is summarized in Figure ~\ref{fig:overall-arch}. While client computation is significantly reduced in SL than FL, this comes at cost of an increase in client-server communication and often in-efficient convergence. \textit{First}, since the client is dependent upon the server for training gradient, required \textit{communication budgets} increase as the client interacts with the server in every iteration of a round (vs once-per-round in FL). This server is also blocked to train synchronously with each client. \textit{Second,} since clients sequentially update shared parameters on the server, convergence may be in-efficient or sub-optimal, especially when client data is heterogeneous. Alleviating these concerns is the focus of this work.

We introduce \textit{AdaSplit}, which enables split learning to scale to low-resource scenarios. \textit{First, } a key insight in AdaSplit is to eliminate client dependence on server gradient which reduces communication cost and also enables asynchronous (and reduced) computation. \textit{Next}, motivated by the fact that neural networks are vastly overparameterized, AdaSplit is able to improve performance by constraining heterogeneous clients to update sparse partitions of the server model. As shown in Figure ~\ref{fig:intro-figure}, this enables AdaSplit to not only achieve improved performance under fixed resources (higher accuracy when similar bandwidth and compute) but also adapt to variable resource budgets (the trade-off curve). Furthermore, to capture and benchmark this multi-dimensional nature of distributed deep learning, we also introduce \textit{C3-Score}, a metric to evaluate performance under resource budgets.


The contributions of this work can be summarized as:
\begin{itemize}
    \item We introduce \textit{AdaSplit}, an architecture for distributed deep learning that can adapt to and improve performance across variable resource constraints.
    \item We introduce \textit{C3-Score}, a metric to benchmark and compare distributed deep learning techniques. 
    \item We validate the effectiveness of \textit{AdaSplit} through experiment comparison with state-of-the-art methods and sensitivity analysis of different design choices.
\end{itemize}


\section{Preliminaries and Motivation}
\label{sec:prelim}
\textit{First}, we formalize the protocol and notation for the split learning (SL) framework. This is also visualized in Figure ~\ref{fig:overall-arch}. For completeness, we also visualize the FL protocol in the same Figure. While the training protocol may appear different, we unify their design choices along key dimensions. \textit{Next}, we formulate three key design dimensions and contextualize specific choices of FL and SL. This helps motivate our proposed AdaSplit technique.

\subsection{Split Learning}
\label{sec:prelim-sl}
Consider a distributed learning setup with $N$ participating clients and $1$ coordinating server. The key idea of split learning (SL) is to distribute (or \textit{split}) the parameters of the training model across client and server. Each client $i$, for $i \in [1,2,...,N]$ is characterized by a local client dataset $D_{i}$, local client model $M_{i}^{c}$ and a single server model $M^{s}$ which is updated by all the clients. The training protocol is executed over $R$ rounds of $T$ iterations each. In each round, the $N$ clients sequentially obtain access to interact with the server for model training over $T$ iterations.  In each iteration $j$ (for $j \in [1,2,,..,T]$) of client $i$ updates the parameters of $M^{s}$ and $M^{c}_{i}$. \textit{First}, a mini-batch ($x_{i}, y_{i}$) is sampled from $D_{i}$ and passed through layers of client model $M^{c}_{i}$ to generate activations $a_{i}$ ($= M^{c}_{i}(x_{i})$). In this document, we may refer to $a_{i}$ as \textit{split activations}. \textit{Second}, the pair of ($a_{i}, y_{i}$) is transmitted to the server. \textit{Third}, at the server, $a_{i}$ is passed through layers of server model $M^{s}$ to generate predictions $\hat{y_{i}}$ ($=M^{s}(a_{i})$). The loss function $L(y_{i}, \hat{y_{i}})$ is computed to generate gradients which are used to locally update parameters of $M^{s}$ and then transmitted to the client to update parameters of $M^{c}_{i}$. In the classical setup, clients follow a round-robin mechanism where client $i+1$ can start interacting with the server only after client $i$ has completed its $T$ iterations for the round. The global model is synchronized implicitly across clients by updating weights of the shared server model $M^{s}$. Furthermore, in some variants, clients models is transmitted between pairs of clients during a round ~\cite{gupta2018distributed} or averaged over all clients after the round ends ~\cite{thapa2020splitfed}. Extensive work has been conducted to establish privacy in split learning and, while beyond scope of this paper, we briefly discuss that in Section ~\ref{sec:related-work}.

\subsection{Design Dimensions: 3C's}
\label{sec:prelim-design-dim}
We define three key design dimensions which focus on how i) model is trained on local client data (\textit{Computation}) and, ii) updates across the clients are synchronized, via the server, into a global model (\textit{Communication} and \textit{Collaboration}).

\textbf{\underline{1. Computation:}} This governs how the processing of data at each client is executed between client and server. Hence, the computation cost can be defined as the total sum of floating-point operations (FLOPs) executed on the client and server. For $N$ clients, this cost ($C1$) can be represented as:
\begin{equation}
    C1 = \sum_{i=1}^{N} R*(F^{c}_{i}*T^{c}_{i} + F^{s}_{i}*T^{s}_{i})
\end{equation}
where, $F^{c}_{i}$ are the FLOPs executed on client for $T^{c}_{i}$ iterations, $F^{s}_{i}$ are FLOPs executed on server for $T^{s}_{i}$ iterations when training with data for client $i$ and $R$ is number of rounds. $F^{c}_{i}$ and $F^{s}_{i}$ increase (or decrease) monotonically with increase (or decrease) in size of client model $M^{c}_{i}$ and server model $M^{s}$ respectively. In federated learning, $F^{s}_{i} = 0$ and $T^{s}_{i} = 0$ since the entire model is executed on client device ($M^{s} = 0$). In contrast, split learning allows to split the model and distribute $F_{c}$ and $F_{s}$ between client and server, based on resource availability. This flexibility of split learning allows scaling to low-resource setups where clients are compute constrained (but servers may scale horizontally) and is a key aspect for design of AdaSplit. However, we also note that this classical split learning framework increases compute load on the server and also blocks the server to train synchronously with each client. 

\textbf{\underline{2. Communication:}} This governs how client-and-server interact with each other. Hence, the communication cost can be defined as the total payload that is transmitted between each of the $N$ client-server pairs over multiple rounds of training. Federated and Split Learning differ based on the modality of the payload and frequency of interaction. However, without loss of generality, this cost ($C2$) can be represented as:
\begin{equation}
\label{eq:comm-prelim}
    C2 = \sum_{i=1}^{N}\sum_{j=1}^{R}\sum_{k=1}^{T} (P_{is} + P_{si})*\sigma(i,j,k)
\end{equation}
where $N$ is number of clients, $R$ is training rounds and $T$ is iterations per round. $P_{is}$ is the payload transmitted from a given client $i$ to server $s$ and $P_{si}$ is the payload transmitted from server $s$ to client $i$. $\sigma(i,j,k)$ denotes if client $i$ interacts with server during iteration $k$ of round $j$. In federated learning, client-server interact using model weights once-per-round. Hence, size of each $P_{is}, P_{si}$ is size of the total model and $\sigma(i,,j,k)= 1$ \textit{only} for $k=T$ (last iteration of every round). In split learning, $P_{is}, P_{si}$ is size of a batch of activations and gradients respectively and $\sigma(i,,j,k)= 1$ $\forall i,j,k$ since client depends upon server for gradient. We note that, even though size of the payload is relatively smaller for split learning (one activation batch vs full model), the high frequency of communication may result in more bandwidth consumption than federated learning.

\textbf{\underline{3. Collaboration:}} This governs how learning (or updates) from local data across the clients is synchronized in the global model. Unlike communication and computation, the cost is non-trivial to define but the impact is measured from the converged accuracy. If the client datasets $D_{i}$ for $i \in [1,2,..,N]$ could be centralized, the unified dataset $D$ ($={D_{1} \cup D_{2} ... \cup D_{N}}$) can be used to train a performant model with gradient descent by sampling iid batches $b \sim D$. Federated and split learning require mechanisms to achieve convergence when this data is decentralized. Abstractly, federated learning executes this by averaging client model parameters (or gradients) on the server after each round, and split learning executes this by requiring all clients to (sequentially) update shared parameters of the server during the round. 

In federated training, the global model in a round $r$ and consequently updated client models ($M^{c}_{i}$) are obtained as:
\begin{equation}
    \label{ref:eqn-client-avg}
    \begin{gathered}
    M^{g} = \sum_{i=1}^{N}(M^{c}_{i}*p_{i}^{r}) \\
    M^{c}_{i} = M^{g}  \forall i \in [1,2,...,N]
    \end{gathered}
\end{equation}
where $p_{i}^{r}$ is a weight assigned to client $i$ in round $r$.

In split training during each round $r$, the server model ($M^{s}$) is updated sequentially by all client $i$ for $\forall i \in [1, 2,...,N]$ as:
\begin{equation}
    \begin{gathered}
            M^{s} =  M^{s} - \alpha*\triangledown \hat{L}( M^{s}(a_{i}), y_{i}) \\
    \end{gathered}
\end{equation}
In some variants of split learning such as ~\cite{thapa2020splitfed}, local client models may also synchronized, at end of each round, similar to federated learning using equation ~\ref{ref:eqn-client-avg}. Then, the global model is obtained by stacking the server and client models. We note that when data across clients are non-iid (that is common on real-world distributed setup), inefficient or sub-optimal converged accuracy is observed in $M^{s,r}$. In split learning, we posit that this happens since gradients from non-iid activations sequentially update the same parameters which violates assumptions of empirical risk minimization ~\cite{vapnik1992principles}.

\section{AdaSplit}
\label{sec:method}
In this section, we delineate the design choices of AdaSplit along each of the three dimensions. We also discuss corresponding trade-offs that enable AdaSplit to adapt to variable resource constraints. Unless specified otherwise, we follow the same notation as defined in Section ~\ref{sec:prelim}. The architecture is visualized in Figure ~\ref{fig:overall-arch}.

\subsection{Computation}
\label{ref:slpp-compute}
The training model is split between the client and server. Following split learning, each client is characterized by a local client model $M^{c}_{i}$ and a global server model $M^{s}$ that is shared across all clients. This flexibility to distribute the model allows scaling split learning (and AdaSplit) to low resource setups. Recall from Section ~\ref{sec:prelim-design-dim}, that in classical split learning, this increases computation load on the server and also blocks the server to train synchronously with each client model which depends upon the server for the gradient.

AdaSplit alleviates these concerns by i) eliminating the dependence of the client model on server gradient and ii) \textit{only} training the server intermittently. This further lowers the total computation cost by decreasing $T_{s}$ (compute iterations on the server) and also unblocks the server to execute asynchronously from the client.

\textbf{Local Client Gradient:} \textit{First}, AdaSplit generates the gradient for training client model on-client itself using a local objective function $L_{client}$ which is a supervised version of NT-Xent Loss \cite{NIPS2016_6b180037}. Given an input batch, $b \sim D_{i}$, then for each input $(x_{i}, y_{i}) \sim b$, $L_{client}$ is applied on a projection ($H(.)$) of the activations $a_{i}$ generated by the client model ($=M^{c}_{i}(x_{i})$). Let $q_i = H(a_i)$ be the corresponding embedding of an input $x_i$, and $Q^{i}_{+}$ be the set of embeddings of other inputs with the same class as $x_i$ in the batch b, the loss can be represented as below:
\begin{equation}
    L_{client} = \sum^{|b|}_{i=0} \sum_{q_{+} \in Q^{i}_{+}} -\log \frac{exp(q_i \cdot q_{+}/\tau)}{\sum^{|b|}_{j \ne i} exp(q_i \cdot q_j/\tau)}
\end{equation}

Here, $\tau$ is a hyperparameter, which controls the "margin" of closeness between embeddings. We set $\tau = 0.07$ in all our experiments.
The pairs (anchor $q_{i}$, positive inputs $q_{+}$) required in $L_{client}$ are sampled using the ground truth labels ($y_{i}$) locally on client.

\textbf{Intermittent Server Training:} \textit{Second}, AdaSplit also \textit{splits} the $R$ round training into two phases: A) \underline{\textit{Local Phase}} B) \underline{\textit{Global Phase}}. \textit{Local Phase} lasts for the first $\kappa$ rounds when only the client model trains, asynchronously and without interacting with the server, using $L_{client}$. After $\kappa$ rounds (till end), the \textit{Global Phase} starts where client interacts with the server by transmitting activations. The server model \textit{only now} start training on data from the clients. The server model $M^{s}$ is optimized using a server loss function ($L_{server}$) which is cross-entropy ($L_{ce}$) for classification tasks. We note that, even in global phase, client model continues to train using $L_{client}$ and \textit{does not} receive any gradient from the server. 

Essentially in AdaSplit, client models leverage compute resources of the server only when required. AdaSplit can adapt to variable computation budgets by regulating two key hyperparameters: i) size of the client model ($\mu$) (for client compute), ii) duration of local phase ($\kappa$) (for server compute). We study the specific impact of these design choices in Section ~\ref{sec:ablations}. Also, in practice, we observe considerable reductions in total computation since $\kappa$ can assume large values (0.8*R), where $R$ is total training rounds, without significant loss of performance. We corroborate this with results in Section ~\ref{sec:results}.


\subsection{Communication}
\label{ref:slpp-comm}
Recall from Section ~\ref{sec:prelim-design-dim} that in classical split learning, the high client-server interaction can be prohibitive for communication cost. AdaSplit alleviates this problem by i) decreasing the payload size and ii) the frequency of communication.

\textbf{Smaller Payload:} \textit{First,} we note that eliminating client dependence on server gradient also significantly reduces communication cost, apart from decreasing computation. In AdaSplit, $P_{si} = 0$ (from equation ~\ref{eq:comm-prelim} in Section ~\ref{sec:prelim-design-dim}) throughout training for each client $i$. Through sensitivity analysis in Section ~\ref{sec:ablations}, we validate that this design choice marginally drops the performance while significantly reducing communication.

\textbf{Infrequent Transmission:} \textit{Second}, we note that two-phase training is also beneficial for communication. In the \textit{Local Phase}, there is no client-to-server communication, with the payload $P_{is} = 0$ for all clients $i$ (from equation ~\ref{eq:comm-prelim} in Section ~\ref{sec:prelim-design-dim}). In the \textit{Global Phase}, clients may start transmitting activations to the server. In this phase, only a subset of clients communicates with the server in each round. Specifically, we introduce an \textit{Orchestrator} (\textit{O}) which resides on the server and uses a running statistic of local client losses to select ($\eta N$) clients in each iteration, that communicate with the server. In AdaSplit,\textit{O} uses a UCB ~\cite{ucb} strategy to prioritize clients who need the server model to improve performance on their data (exploitation) while also ensuring that the final model can generalize well to different client data distributions (exploration).

Let $S^{t}_{i}$ is a binary flag denoting if client $i$ is selected at iteration $t$ and $L_{i}^{t}$ denote the server loss from activations ($a_{i}$) for the iteration. At each iteration $t$, selected clients (i.e. $S_{i}^{t}=1$) transmit input activations to update server model and the loss $L_{i}^{t}$ is stored. For unselected clients (i.e. $S_{i}^{t} = 0$), $L_{i}^{t}$ is defined the average of their loss value in previous iterations ($L_{i}^{t} = \frac{L^{t-1}_{i} + L^{t-2}_{i}}{2}$). Here, we note that $L_{i}^{t}$ is only used for selection and the client model continues to train locally with $L_{client}$. \textit{Finally,} \textit{O} assigns a new score to each client using the advantage function and clients with the top-$\eta$ scores are selected for next iteration. The advantage function ($A_{i}$) for ~\cite{ucb} is defined below:
\begin{equation}
    \begin{gathered}
        A_i = \frac{l_i}{s_i} + \sqrt{\frac{2 \log T}{s_i}} \\
    \end{gathered}
\end{equation}
where, $l_i = \sum^{T}_{t=0}\gamma^{T-1-t} \cdot L^{t}_{i}$, $s_i = \sum^{T}_{t=0}\gamma^{T-1-t} \cdot S^{t}_{i}$ and $T$ is total iterations in the round. $\gamma \in [0, 1]$ is a hyperparameter that determines the importance of historical losses. We initialize $L^{t}_i = 100$ for all clients for $t=0$ and $t=1$. 

Before proceeding, we make a few statements here. \textit{First,} we note that subset selection has previously been used in FL to regulate communication cost ~\cite{fedavg, fedprox, gaurijoshi-fl} where the global model after a round may be obtained from few clients only (see variable $p_{i}^{r}$ in equation ~\ref{ref:eqn-client-avg}). \textit{However}, classical split learning does not have a similar infrastructure since each client is entirely dependent on the server for gradient during training. Eliminating client dependence on server gradient in AdaSplit helps realize the benefit. \textit{Finally}, we highlight that the design of orchestrator is specialized for AdaSplit where it needs to be invoked in each iteration (vs rounds in FL) and selects client for training (vs model averaging in FL).

AdaSplit can adapt to variable communication budgets by regulating two key hyperparameters: i) the number of selected clients ($\eta$), ii) the duration of the local phase ($\kappa$). We study the specific impact of these design choices in Section ~\ref{sec:ablations}. In practice, we observe considerable reductions in communication cost since $\kappa$, $\eta$ can assume large values ($\kappa = 0.8*R, \eta = 0.6*N$) without significant loss of performance. We corroborate this with results in Section ~\ref{sec:results}.

\subsection{Collaboration}
\label{ref:slpp-collab}
AdaSplit, like split learning, synchronizes updates in the global model by requiring clients to sequentially update shared server model parameters.
Recall from Section ~\ref{sec:prelim-design-dim} that when inter-client data is heterogenous, this often results in the global model converging to sub-optimal accuracy. To alleviate this, the key idea in AdaSplit is to have each client update only a partition of the server model ($M_{s}$) parameters. The motivating insight is to take advantage of the fact that neural network models are vastly over-parameterized ~\cite{cl-vnp-14} and only a small proportion of the parameters can learn each (client's) task with little loss in performance~\cite{cl-vnp-13, cl-vnp, cl-vnp-11, cl-vnp-18}. 

\textbf{Update Sparse Partitions of Server Model:} 
During the \textit{global phase}, we add an $L^{1}$ weight regulator to promote sparsity in the server model $M^{s}$. Specifically, instead of pruning the network, we learn a client ($i$) specific multiplicative mask $m_{i}$ which constrains the set of $M^{s}$ parameters client $i$ can update. Given batch of activations $a_{i}$ from client $i$, server model $M^{s}$ is updated as:
\begin{equation}
    M^{s} = M^{s} - \alpha*\hat{m}_{i}*\triangledown \hat{L}( M^{s}(a_{i}), y_{i})
\end{equation}
This simulates relative sparsity (for each client) in $M^{s}$ without pruning any parameters since goal is to increase server model capacity (to accommodate many diverse clients) rather than achieving compression. Here, $m_{i}$ evolves during training and is forced to be extremely sparse by optimizing the following objective on the server for each client $i$:
\begin{equation}
    L_{server} = L_{ce}(\hat{y_{i}}, y_{i}) + \lambda*\omega(m_{i})
\end{equation}
where, $\omega(.)$ is an $L^{1}$ regularizer and $\hat{y_{i}} = M^{s}(M^{c}_{i}(x_{i}))$. The $\lambda$ hyperparameter promotes sparsity of the masks and can be intuitively visualized as controlling the extent of collaboration between clients on the server. At inference, the effective server model for client is $M^{s}*m_{i}$ where $m_{i}$ is a highly sparse binary mask and may be stored on client. Results in Section ~\ref{sec:results} show that this strategy of regulating collaboration significantly improves performance. Finally, we note similarities between \textit{each round} of collaboration in AdaSplit and continual learning, albeit AdaSplit works in activation space and is iterative. However, we anticipate exploring this connection may present interesting directions of future work.

\section{Experimental Setup}
\textit{First,} we formalize the experimental protocol with datasets and baselines. \textit{Next,} we define the evaluation protocol and introduce the \textit{C3-Score} as a unified metric to benchmark and compare the efficiency of distributed deep learning techniques. \textit{Finally}, we summarize implementation details for results presented in this work.

\subsection{Datasets}
To robustly validate the efficacy of AdaSplit, we conduct extensive experiments on benchmark datasets and simulate varying levels of inter-client heterogeneity. Specifically, we design two experimental protocols, as described next: \textbf{a) Mixed-CIFAR:} We divide the 10 classes of CIFAR-10 ~\cite{cifar} into 5 subsets of 2 distinct classes each. Every client is assigned data from one of the 5 subsets. In this protocol, there is low and consistent heterogeneity between data across all pairs of clients. \textbf{b) Mixed-NonIID:} We use 5 benchmark datasets: i) MNIST \cite{mnist} ii) CIFAR-10 ~\cite{cifar} iii) FMNIST ~\cite{fmnist} iv) CIFAR-100 ~\cite{cifar} v) Not-MNIST ~\cite{not-mnist} and each client receives samples from exactly one dataset. In this protocol, there is high and variable inter-client heterogeneity between client pairs. For instance, clients with FMNIST and MNIST have lower pairwise-heterogeneity between each other and high pairwise heterogeneity with clients containing CIFAR-100. For experiments with both protocols, input images are scaled to 32x32x3 and grayscale MNIST/FMNIST images are transformed by stacking along channels. We will release the evaluation protocol with the code for use by the research community.

\subsection{Baselines}
We compare state-of-the-art split learning and federated learning techniques. Specifically, for split learning, we compare with SL-basic ~\cite{gupta2018distributed} and SplitFed ~\cite{thapa2020splitfed}. To ensure validity of analysis and also highlight efficacy of results, we also compare with state-of-the-art federated learning techniques: FedAvg ~\cite{fedavg}, FedNova ~\cite{fednova}, Scaffold ~\cite{scaffold} and FedProx ~\cite{fedprox}. These techniques are specially designed for heterogenous (non-iid) setups and provide strong benchmarking for the efficacy of AdaSplit.

\begin{table*}[]
\centering
\tabcolsep=0.45cm
\begin{tabular}{l|c|c|c||c}
\toprule
Method         & Accuracy & Bandwidth (GB) & Compute (TFLOPS) & \textbf{C3-Score} \\
\midrule
SL-basic ~\cite{gupta2018distributed} &   $84.65 \pm 0.32$        &     84.54      &  3.76 (15.14)  & 0.72   \\
SplitFed ~\cite{thapa2020splitfed}      &    $84.67 \pm 0.24$     &      84.64     &     3.76 (15.14)  & 0.73  \\
\midrule
FedAvg ~\cite{fedavg} &     $82.21 \pm 0.19$     &      2.39     &  17.13 (17.13)  & 0.72 \\
FedProx ~\cite{fedprox} &    \textbf{85.09 $\pm$ 0.29}      &     2.39     &    17.13 (17.13)  &  0.75     \\
Scaffold ~\cite{scaffold}  &     $84.73 \pm 0.17$       &     4.78      &    17.13 (17.13)    & 0.74 \\
FedNova ~\cite{fednova}       &    $82.71 \pm 0.27$      &     2.39      &   17.13 (17.13)  & 0.73 \\
\midrule
\textbf{AdaSplit ($\kappa$=0.6, $\eta$=0.6)}          &    \textbf{88.88 $\pm$ 0.27}      &      9.71     &        5.38 (8.82) & \textbf{0.85} \\
\textbf{AdaSplit($\kappa$=0.75, $\eta$=0.6)}          &    \textbf{87.11 $\pm$ 0.59}      &      2.43     &        5.38 (10.88) & \textbf{0.83} \\
\bottomrule  
\end{tabular}
\caption{Results on \textbf{Mixed-NonIID} dataset. AdaSplit achieves improved performance while reducing resource (bandwidth, compute) consumption. This is corroborated by the \textit{C3-Score} (higher is better).\vspace{-0.7cm}}
\label{tab:non-iid}
\end{table*}

\begin{table*}[]
\centering
\tabcolsep=0.45cm
\begin{tabular}{l|c|c|c||c}
\toprule
Method         & Accuracy & Bandwidth (GB) & Compute (TFLOPS) & \textbf{C3-Score} \\
\midrule
SL-basic ~\cite{gupta2018distributed} &   $67.90 \pm 3.52$        &     34.88      &  1.66 (13.76)      & 0.59  \\
SplitFed ~\cite{thapa2020splitfed}      &    $71.46 \pm 2.13$      &      35.94     &    1.66 (13.76)    & 0.62 \\
\midrule
FedAvg ~\cite{fedavg} &    $91.31 \pm 0.49$      &      2.39     &  11.77 (11.77)      & 0.79  \\
FedProx ~\cite{fedprox} &    \textbf{92.54 $\pm$ 0.48}      &       2.39    &   11.77 (11.77)  & 0.81 \\
Scaffold ~\cite{scaffold} &     $87.30 \pm 1.36$       &     4.79      &    11.77 (11.77)    & 0.76 \\
FedNova ~\cite{fednova}       &    $88.94 \pm 0.32$      &     2.39      &   11.77 (11.77)   & 0.77    \\
\midrule
\textbf{AdaSplit ($\kappa$=0.6, $\eta$=0.6)}          &    \textbf{91.92 $\pm$ 1.88}      &      2.85     &        2.38 (4.81)  & \textbf{0.89} \\
\textbf{AdaSplit ($\kappa$=0.3, $\eta$=0.6)}          &    \textbf{92.91 $\pm$ 0.91}      &      6.57     &        2.38 (6,63)  & \textbf{0.88}  \\
\bottomrule
\end{tabular}
\caption{Results on \textbf{Mixed-CIFAR} dataset. AdaSplit achieves improved performance while reducing resource (bandwidth, compute) consumption. This is corroborated by the \textit{C3-Score} (higher is better).\vspace{-0.8cm}}
\label{tab:split-cifar}
\end{table*}

\subsection{Evaluation Metrics: C3-Score}
We evaluate performance both independently along each of the dimensions as well as using a unified metric. To evaluate along design dimensions, we report \textit{Accuracy}, \textit{Bandwidth}  and \textit{Compute}. Accuracy is measured as mean and standard deviation over multiple independent runs with different seeds. \textit{Bandwidth} is measured in GB and \textit{Compute} is measured in TFLOPS. We note that, in real-world cases, servers may scale horizontally and bottleneck is often at client. For completeness, we report both client compute and total (client+server) compute.

\textbf{C3-Score:} For an efficient method, the goal is to maximize accuracy while minimizing resource (bandwidth, compute) consumption. We introduce \textit{\textbf{C3-score}}, a metric to evaluate joint performance of any distributed deep learning technique in this multi-dimensional setup. Let, $B_{max}, C_{max}$ be the maximum resource budgets for bandwidth and client compute. We note that $A_{max} = 100\%$ (max achievable accuracy) for predictive tasks. Consider, a given method $m$, with mean accuracy $A_{m}$, bandwidth consumption $B_{m}$ and client compute consumption $C_{m}$. Then, the \textit{C3-Score} is defined as below:
\begin{equation}
    C3-Score(A_{m}, B_{m}, C_{m}) = (\hat{A}_{m})*e^{-( \hat{B}_{m} + \hat{C}_{m} )/T}
\end{equation}
where, $\hat{B}_{m} = B_{m} / B_{max}, \hat{C}_{m} = C_{m} / C_{max}, \hat{A}_{m} = A_{m} / A_{max}$ and $T$ is a scaling temperature.

The \textit{C3-Score} always lies between 0 and 1 with a higher score representing a better (more efficient) method. Trivially, C3-Score $\to 0$ as $\hat{B}_{m} \to \infty$ or $\hat{C}_{m} \to \infty$ (as consumption increases or budget decreases). Here, we make a few comments about the metric. \textit{First,} considering resource budgets is useful for real-world use and can also enable to directly eliminate methods that exceed the budget. We take motivation from work in differential privacy (privacy budgets) which contextualize comparing between methods. Also, normalizing cost with the budget allows realizing a bounded metric which is useful for comparison and benchmarking. Finally, we assume a multiplicative form to the metric to enable an easy extension to other resource dimensions in the future. Specifically, incorporating privacy budgets is an interesting direction of future work.

\subsection{Implementation Details}
All experiments are trained for (R=20) rounds with 1 epoch per round using the same convolutional (LeNet) backbone. Results are reported for 5 (=N) clients. For the federated learning baselines, we use open-source implementations provided in ~\cite{niid-bench}. For robust comparison, we also tuned parameters for these baselines and note some performance gain was observed (over default values) which is then used for comparison. For all split learning baselines, we set the default client model size is 20\% ($\mu = 0.2$) and use Adam optimizer with a learning rate of 1e-3. This same optimizer configuration is used for both client and server loss in AdaSplit. In AdaSplit, the default parameters are: a) $\kappa$ = 0.6, $\eta$ = 0.6, $\gamma$ = 0.87, $\lambda$ = 1e-5 (for Mixed-CIFAR) and 1e-3 (for Mixed-NonIID). For experiments on \textbf{Mixed-CIFAR} and \textbf{Mixed-NonIID}, the communication and computation budgets are chosen to be the performance of worst-performing baselines on the corresponding dataset respectively. Experiments are implemented in Pytorch, executed on 1 NVIDIA RTX-3060 GPU and reported over 5 independent runs.

\section{Results}
\label{sec:results}
We report performance on ~\textbf{Mixed-CIFAR} in Table ~\ref{tab:split-cifar} and  ~\textbf{Mixed-NonIID} in Table ~\ref{tab:non-iid}. For \textit{C3-Score}, we set the bandwidth and communication budgets to be $B_{max} = 35.94$ GB and $C_{max} = 11.77$ TLFOPS on \textit{Mixed-CIFAR} and $B_{max} = 84.64$ GB and $C_{max} = 17.13$ TFLOPS on \textit{Mixed-NonIID}. These values correspond to the highest bandwidth and computation cost among all the methods for the corresponding datasets. The results on both datasets consistently support the following key observations:

\circled{1} \textbf{AdaSplit} \textit{outperforms other split learning techniques} and achieves significantly better accuracy while also reducing bandwidth consumption. For instance, on \textit{Mixed-CIFAR} (Table ~\ref{tab:split-cifar}), in comparison to SL-basic, AdaSplit \textbf{improves performance by 24\%} and consumes \textbf{89\% lower} bandwidth. This is corroborated by an increase in C3-Score from 0.59 for SL-basic ~\cite{gupta2018distributed} to 0.89 for AdaSplit. Furthermore, similar trend is observed on \textit{Mixed-NonIID} (Table ~\ref{tab:non-iid}) as well. Specifically, AdaSplit achieves accuracy of 88.88 against 84.67 for SplitFed while consuming \textit{75 GB less bandwidth}. The corresponding trend is also captured by \textit{C3-Score} which is 0.85 for AdaSplit against 0.73 for SplitFed ~\cite{thapa2020splitfed}.

\circled{2} \textbf{AdaSplit} \textit{makes} \textit{split learning a competitive alternative to federated learning}. On both datasets, we observe that AdaSplit consistently achieves higher (or similar) accuracy with significantly lower client compute and similar bandwidth. For instance, on \textit{Mixed-NonIID}, AdaSplit achieves 87.11\% accuracy with 2.43 GB bandwidth and 5.38 TFLOPS compute. In comparison, the closest federated learning baseline, FedProx, achieves 85\% accuracy but consumes 17.13 TFLOPS (3x more than AdaSplit) and with similar bandwidth (2.39 GB). This is corroborated with an improved C3-Score of 0.85 for AdaSplit and 0.75 for FedProx.

\circled{3} \textbf{AdaSplit} \textit{consistently provides the \textbf{best trade-off among all} of federated and split learning baselines}. For instance, on \textit{Mixed-CIFAR}, AdaSplit achieves a \textit{C3-Score} of 0.89 with the closest federated learning baseline (FedProx) ~\cite{fedprox} is at 0.81, FedAvg ~\cite{fedavg} at 0.79 and SplitFed at 0.62. Furthermore, similar trend is observed on \textit{Mixed-NonIID} as well. Specifically, AdaSplit achieves a \textit{C3-Score} of 0.85 with the closest baseline FedProx at 0.75, Scaffold ~\cite{scaffold} at 0.74 and SL-basic ~\cite{gupta2018distributed} at 0.72. 

\circled{4} \textbf{AdaSplit} \textit{can adapt to variable resource budgets}. From results on \textbf{Mixed-NonIID} (Table ~\ref{tab:gradient-flow}), we can see that given a higher communication budget (13.36 GB), AdaSplit can further improve accuracy to 89.77\% which corresponds to a 5\% improvement over FedProx ~\cite{fedprox}. Figure ~\ref{fig:intro-figure} shows trade-offs that AdaSplit can achieve for accuracy by varying bandwidth (and compute) budget. We note that the respective trade-off curves for bandwidth (and compute) are obtained while keeping compute (and bandwidth) budgets fixed respectively. We discuss in more detail in Section ~\ref{sec:ablations}.

\section{Discussion}
\label{sec:ablations}
In this section, we conduct a sensitivity analysis of key design choices in AdaSplit and analyze the consequent impact on accuracy and resource consumption. Results validate the ability of AdaSplit to efficiently adapt to variable resource budgets. Unless specified otherwise, the hyperparameters used are $\kappa=0.6$, $\eta=0.6$, $\mu=0.2$.

\circled{1} \textbf{Varying Size of Client Model:}
Table ~\ref{tab:client-size-cifar} presents results from varying number of layers on client for \textit{Mixed-CIFAR10} dataset. We observe that \textit{Computation} on client increases monotonically with the number of client layers. We also observe a decrease in \textit{Communication} cost as evident from lower bandwidth. This can be attributed to the convolution design of the model where \textit{split activations} becomes smaller with depth (reducing payload $P_{is}$). Also, we note marginal gain in performance for larger server model since it provides more parameters for \textit{Collaboration}. We observe similar trends on \textit{Mixed-NonIID} and include results in the appendix. Hence, \textit{AdaSplit adapts to variable client computation budgets}.

\begin{table}[]
\centering
\tabcolsep=0.19cm
\begin{tabular}{l|c|c|c}
\toprule
$\mu$ & Accuracy                      & Bandwidth (GB) & Compute (TFLOPS)      \\ 
\midrule
0.2         & 91.92 $\pm$ 1.88 & 2.85      & 2.38 (4.81)   \\
0.4         & 92.12 $\pm$ 1.61 & 1.18      & 9.04 (9.85)   \\
0.6         & 86.37 $\pm$ 6.74 & 1.08      & 11.58 (11.68) \\
0.8         & 90.14 $\pm$ 2.80 & 1.05      & 11.95 (11.97) \\ 
\bottomrule
\end{tabular}
\caption{Results on \textit{Mixed-CIFAR10}. Varying number of client layers ($\mu$) enables AdaSplit to adapt to variable client computation budgets. \vspace{-0.8cm}}
\label{tab:client-size-cifar}
\end{table}

\circled{2} \textbf{Varying Duration of Local Phase:}
Table ~\ref{tab:interrupt-duration} presents results from varying $\kappa$ on \textit{Mixed-CIFAR10} dataset. We observe that \textit{Communication} cost decreases as $k$ increases. This is because $P_{is} = 0$ for all rounds $r < \kappa$ on given client $i$. \textit{Computation} cost of the server also decreases on increasing $\kappa$ though client compute is unchanged. note marginal decrease in accuracy since higher $\kappa$ allows for fewer \textit{collaboration} iterations on server. Specifically, increasing $\kappa$ from $0.3$ to $0.9$ decrease accuracy from 89.80\% to 87.11\%, while bandwidth falls drastically from $17.22$ GB to $2.43$ GB. This is corroborated on \textit{Mixed-NonIID} as shown in Table ~\ref{tab:gradient-flow}. Hence, \textit{AdaSplit adapts to variable communication and server computation budgets}.
\begin{table}[b]
\centering
\tabcolsep=0.19cm
\begin{tabular}{l|c|c|c}
\toprule
$\kappa$ & Accuracy  & Bandwidth (GB) & Compute (TFLOPS)   \\ 
\midrule
0.3         & 92.91 $\pm$ 0.91 & 6.57      & 2.38 (6.63) \\
0.45            & 90.97 $\pm$ 1.02 & 4.72      & 2.38 (5.72) \\
0.6             & 89.77 $\pm$ 1.62 & 3.56      & 2.38 (4.81) \\
0.75            & 88.62 $\pm$ 3.68 & 2.15      & 2.38 (3.90) \\
0.90              & 88.02 $\pm$ 0.91  & 0.89       & 2.38 (2.98) \\
\bottomrule
\end{tabular}
\caption{ Results on \textit{Mixed-CIFAR10}. Varying duration of local phase ($\mu$) enables AdaSplit to adapt to variable communication and server computation budget.\vspace{-0.5cm}}
\label{tab:interrupt-duration}
\end{table}

\circled{3} \textbf{Eliminating Gradient Dependence:}
Table ~\ref{tab:gradient-flow} studies the impact of training client model without gradient from server on \textit{Mixed-CIFAR10} dataset. We observe \textit{Communication} cost decreases significantly with bandwidth reduced by one-half. We observe accuracy is generally insensitive though there is slight increase in standard deviation. Hence, \textit{AdaSplit adapts to variable communication budget} and provides consistent performance. 
\begin{table}[t]
\begin{tabular}{l|c|c}
\toprule
$\kappa$                 & Accuracy      & Bandwidth (GB) \\
\midrule
\multirow{2}{*}{0.3}  & 89.80 $\pm$ 0.38 & 17.22     \\
                      & 89.96 $\pm$ 0.23 & 34.84     \\
\midrule
\multirow{2}{*}{0.45} & 89.77 $\pm$ 0.34 & 13.36     \\
                      & 89.47 $\pm$ 0.21 & 27.18     \\
\midrule
\multirow{2}{*}{0.60} & 89.08 $\pm$ 0.38 & 9.65      \\
                      & 89.03 $\pm$ 0.28 & 19.79     \\
\midrule
0.75                  &  88.17 $\pm$ 0.59 & 6.10      \\
                      &  88.31 $\pm$ 0.40 & 12.06     \\
\midrule
\multirow{2}{*}{0.90}     & 87.11 $\pm$ 0.45 & 2.43      \\
                     & 87.05 $\pm$ 0.39 & 4.89     \\
\toprule
\end{tabular}
\caption{Results on \textit{Mixed-NonIID}. In each Accuracy cell, Row-1 trains client with $L_{client}$ and Row-2 trains client with $L_{client} + L_{server}$. Accuracy is largely insensitive to server gradient across various $\kappa$} \vspace{-0.8cm}
\label{tab:gradient-flow}
\end{table}

\circled{4} \textbf{Further Reducing Payload Size:}
While we sparsify server model parameters to improve collaboration in \textit{AdaSplit}, here we also consider sparsification of split activations to reduce communication payload. Specifically, we train the client model with an additional $L^{1}$ regularizer that regulates magnitude of split activations. Results are presented on \textit{Mixed-NonIID} in Table  ~\ref{tab:activation-sparsity}. \textit{Computation} remains unchanged. \textit{Communication} decreases as payload ($P_{ij}$) becomes sparse. This results in fall in accuracy which highlights worsening collaboration. For instance, AdaSplit can train with only 0.76 GB of bandwidth and achieve 85.79\% accuracy. From Table ~\ref{tab:non-iid}, FedProx achieves 85.09\% and consumes 2.39 GB budget. Hence, \textit{AdaSplit adapts to extremely low communication budgets}.

\begin{table}[b]
\centering
\tabcolsep=0.19cm
\begin{tabular}{l|c|c}
\toprule
$\beta$ & Accuracy      & Bandwidth (GB) \\
\midrule
0      & 91.09 $\pm$ 1.48  & 3.45      \\
1e-7   & 90.52 $\pm$ 2.16  & 3.25      \\
1e-6   & 91.92 $\pm$ 1.89  & 2.85      \\
5e-6   & 87.6 $\pm$ 4.82   & 1.19      \\
1e-5   & 85.79 $\pm$ 4.10  & 0.76      \\
0.0001 & 79.18 $\pm$ 4.81  & 0.08      \\
0.1    & 51.00 $\pm$ 0.42  & 0.0044   \\
\bottomrule
\end{tabular}
\caption{Results on \textit{Mixed-CIFAR10} dataset. Sparsification of split activations enables AdaSplit to adapt to extremely low communication budgets.\vspace{-1cm}}
\label{tab:activation-sparsity}
\end{table}



\section{Related Work}
\label{sec:related-work}
In this section, we review the literature in distributed and collaborative deep learning broadly, and specific research in split learning.
\paragraph{\textbf{Parallelizable Deep Learning}}
For centralized machine learning, some parallelization methods have been proposed to enable training on large-scale data sources. 
\textbf{Data Parallelism} ~\citep{hillis1986data} based distributed ML simulates large mini-batch training by splitting data among multiple identical models and training each model on a shard of the data independently. The key challenge is to ensure consistency of the global model by synchronizing across multiple model copies. This is achieved via i) \textit{Synchronous Optimization} - which synchronizes after every minibatch ~\cite{das2016distributed, chen2016revisiting}, resulting in high communication overhead, and ii) \textit{Asynchronous Optimization} which does global synchronization less frequently ~\cite{dean2012large}, improving communication but at a cost of poor sample efficiency due to staleness of model copies. \textbf{Model Parallelism} splits the model parameters over multiple processors either to improve parallelization~\cite{ben2019demystifying} or when it is too large to fit in the memory of a single device ~\cite{shazeer2018mesh, harlap2018pipedream, lepikhin2020gshard}. This is increasingly common as the size of the state-of-the-art models is exploding (to billions of parameters ~\cite{brown2020language, radford2019language}), which can be challenging for data parallelism. However, model parallelism has its challenges including i) \textit{high communication costs}, due to activations communicated between devices for each mini-batch and ii) \textit{device under-utilization}, due to inter-device control-flow dependency during forward and backward propagation~\cite{ben2018torsten}. \textbf{Local Parallelism} alleviates the staleness issue by eliminating the control flow dependency between devices. The key idea is to perform training using intermediate local objectives. Various local learning configurations are proposed where a local objective can be used greedily for each layer ~\cite{lowe-2019,belilovsky-2020}, for a set of layers on each device ~\cite{laskin2020parallel} or layers across multiple devices ~\cite{gomez2020interlocking}. From a comparison standpoint, federated learning setup is close to data parallelism, split learning extends model parallelism to data-parallel setups and the proposed AdaSplit extends local parallelism to data-parallel setups.

\paragraph{\textbf{Distributed Deep Learning}}
This broadly involves training deep networks with data that is distributed across multiple devices (or clients) and a coordinating server. The two main paradigms for collaborative learning are Federated Learning (FL) ~\cite{fed-avg,kairouz2019advances} and Split Learning (SL) ~\cite{gupta2018distributed,poirot2019split}. This work is focused on the split learning paradigm, with a particular interest in mechanisms to improve the efficiency of communication and collaboration. However, here, we contextualize relevant literature in both federated and split learning along the key design choices introduced in Sec~\ref{sec:prelim}.
\textbf{D1: Computation}
FL generalizes data parallelism (to each client) and the entire model trained locally on-device at the data-owning client. While the computation at the server involves averaging of model parameters, computation at the client-side requires training ML model at every round. This can be a bottleneck for clients that are limited by computation. Some recent works have attempted to address this problem by training on a subset of parameters~\cite{diao2020heterofl} or performing pruning~\cite{li2022hermes,zhou2021adaptcl} on the trained model. All such methods are limited by the extent of pruning that can be performed and require the full model to be trained on the device. In contrast, SL distributes the model parameters between clients and servers, this allows reduced as well as tunable computation load distribution between client and server. Several architectures for parameter distribution are discussed in ~\cite{splitlearning2}. Our proposed AdaSplit allows to further improves computation by eliminating gradient flow and training server intermittently.
\textbf{D2: Communication} In FL, client and server communicate once every training round and this is executed through model weights or gradients of the locally trained model copies. 
This communication cost scales with the size of the model and the number of clients in the system, and is a major challenge for deploying in resource-constrained setups. Various methods have been proposed to reduce this cost through model compression on client~\cite{konevcny2016federated,malekijoo2021fedzip,hamer2020fedboost}, client subset selection~\cite{cho2020bandit,nishio2019client,balakrishnan2020diverse} as well as greedy federated training of client models ~\cite{nishio2019client,mo2021ppfl}. In SL, the client and server communicate in every training iteration using minibatch activations. While each payload is small (and relatively independent of model size), a high frequency of communication is a challenge. In this work, locally-parallel AdaSplit reduces communication costs by alleviating the need for high-frequency server-to-client gradient flow.
\textbf{D3: Collaboration} For Federated Learning, while FedAvg ~\cite{fedavg} performs poorly with non-iid clients, various techniques have been proposed for working in such heterogenous setups ~\cite{zhu2021federated, li2019fedmd, zhao2018federated}. Similarly, SL-basic and SplitFed perform poorly in non-iid setups, resulting in worsened communication cost and accuracy. To the best of our knowledge, no method has been proposed to tackle this issue. AdaSplit alleviates the same through parameter sparsification on the server.

\paragraph{\textbf{Split Learning: Research and Applications}}
While most of the recent works~\citep{kairouz2019advances} in federated learning try to address the aforementioned D1, D2, and D3 aspects, the clients are still restricted by the requirement of executing the complete computation graph on the device. This restriction is circumvented by split learning~\citep{gupta2018distributed} that revamps the federated learning architecture by operationalizing model parallelism for multi-institution collaborative learning. In split learning, clients share activations produced by a neural network instead of the weight or gradients as done in federated learning. Different architectures for forward and backward propagation are illustrated in \cite{vepakomma2018split}. This architecture adds flexibility to the client-side computation by distributing it with the centralized server. A rigorous comparison between FL and SL have been made across several dimensions by \cite{thapa2021advancements} and other aspects have been explored such as communication~\citep{singh2019detailed}, healthcare applications~\citep{gawali2021comparison}, privacy~\citep{kim2020multiple}, IOT~\citep{gao2020end} and frameworks~\citep{vepakomma2018no}. Several recent works have integrated federated and split learning architecture such as SplitFed~\citep{thapa2020splitfed}, SplitFedV3~\citep{gawali2021comparison} and FedSL~\citep{abedi2020fedsl}. In addition to the computational and communication benefit, split learning allows distributed and privacy-preserving prediction that is not possible under the federated learning framework. Consequently, several works have used split learning for inference to build defense~\citep{maxentropy, shredder, osia2020hybrid, xiang2019interpretable, vepakomma2020nopeek, bertran2019adversarially, liu2019privacy, li2019deepobfuscator, singh2021disco, samragh2020private} and attack mechanisms~\citep{he2019model, pasquini2020unleashing, madaan2021vulnerability}. Some of these benefits have led to applied evaluation of split learning for mobile phones~\citep{palanisamy2021spliteasy}, IoT~\citep{park2021communication, koda2020distributed, koda2020communication}, model selection~\citep{sharma2019expertmatcher1,sharma2019expertmatcher2} and healthcare~\citep{poirot2019split}.

\section{Conclusion}
We introduce \textit{AdaSplit}, a technique for scaling distributed deep learning to low resource scenarios. \textit{AdaSplit} builds upon the split learning framework and reduces a) computation by eliminating client dependence on server gradient and training the server intermittently, b) bandwidth by reducing payload size and communication frequency between client-and-server, and c) improves performance by constraining each client to update sparse partitions of server model. To capture and benchmark this multi dimensional nature of distributed deep learning, we also introduce\textit{C3-Score}, a metric to evaluate performance under resource budgets. We validate the effectiveness of \textit{AdaSplit} under limited resources through extensive experimental comparison with strong federated and split learning baselines. We also present sensitivity analysis of key design choices in AdaSplit which validates the ability of \textit{AdaSplit} to provide adaptive trade-offs across variable resource budgets.

\bibliographystyle{ACM-Reference-Format}
\bibliography{main}

\end{document}